# Ascle: A Python Natural Language Processing Toolkit for Medical Text Generation

Rui Yang BS[1,2,*], Qingcheng Zeng BS[3,*], Keen You BS[4,*], Yujie Qiao MS[5,*], Lucas Huang BS[4], Chia-Chun Hsieh BS[4], Benjamin Rosand BS[4], Jeremy Goldwasser BS[4], Amisha D Dave MD[6], Tiarnan D.L. Keenan BM BCh, PhD[7], Emily Y Chew MD[7], Dragomir Radev PhD[4], Zhiyong Lu PhD[8], Hua Xu PhD[9], Qingyu Chen PhD[8,9], Irene Li PhD[10]

1 Centre for Quantitative Medicine, Duke-NUS Medical School, Singapore, Singapore
2 Department of Biomedical Informatics, Yong Loo Lin School of Medicine, National University of Singapore, Singapore, Singapore
3 Department of Linguistics, Northwestern University, Evanston, IL, USA
4 Department of Computer Science, Yale University, New Haven, CT, USA
5 Yale School of Public Health, Yale University, New Haven, CT, USA
6 Yale New Haven Hospital, Yale School of Medicine, Yale University, New Haven, CT, USA
7 Division of Epidemiology and Clinical Applications, National Eye Institute, National Institutes of Health, Bethesda, MD, USA
8 National Center for Biotechnology Information, National Library of Medicine, National Institutes of Health, Bethesda, MD, USA
9 Section of Biomedical Informatics and Data Science, Yale School of Medicine, Yale University, New Haven, CT, USA
10 Information Technology Center, University of Tokyo, Tokyo, Japan

**ABSTRACT**

**Objective**

This study introduces Ascle, a pioneering natural language processing (NLP) toolkit designed for medical text generation. Ascle is tailored for biomedical researchers and healthcare professionals with an easy-to-use, all-in-one solution that requires minimal programming expertise. For the first time, Ascle evaluates and provides interfaces for the latest pre-trained language models, encompassing four advanced and challenging generative functions: question-answering, text summarization, text simplification, and machine translation. In addition, Ascle integrates 12 essential NLP functions, along with query and search capabilities for clinical databases.

**Materials and Methods**

We fine-tuned 32 domain-specific language models and evaluated them thoroughly on 24 established benchmarks. Additionally, for the question-answering task, we conducted manual reviews with clinicians, focusing on Readability, Relevancy, Accuracy, and Completeness, to provide users with a more reliable evaluation.

**Results**

The fine-tuned models consistently improved text generation tasks. For instance, it improved the machine translation task by 20.27 in terms of BLEU score. For the answer generation task, manual reviews showed the generated answers had average scores of 4.95 (out of 5), 4.43, 3.9, and 3.31 in Readability, Relevancy, Accuracy, and Completeness, respectively.

**Conclusions**

This study introduces the development and evaluation of Ascle, a user-friendly NLP toolkit designed for medical text generation. Ascle offers an all-in-one solution including four advanced generative functions: question-answering, text summarization, text simplification, and machine translation. The toolkit, its models, and associated data are publicly available via https://github.com/Yale-LILY/Ascle.

**Keywords:** natural language processing, machine learning, generative artificial intelligence, healthcare

## INTRODUCTION

Medical texts present significant domain-specific challenges, including issues such as ambiguities, frequent abbreviations, the presence of negations, and complexities in segmentation.[1,2] The manual curation of these texts is a time-consuming and labor-intensive process.[3] In response to these challenges, natural language processing (NLP) algorithms have been developed to automate text processing.[2,4,5] Recent years have seen a notable shift towards the use of domain-specific pre-trained language models, transitioning from shallow embeddings like BioWordVec[6] and BioSentVec[7] to advanced architectures like Bidirectional Encoder Representations from Transformers (BERT)[8] such as BioBERT,[9] ClinicalBERT,[10] and PubMedBERT.[11] Furthermore, medical large language models (LLMs) such as Med-PaLM 2[12] have demonstrated powerful generative capabilities, possessing exceptional zero- and few-shot performance. Those domain-specific language models collectively have substantially enhanced the effectiveness of NLP tasks in the biomedical and clinical domains including text classification, named entity recognition, text segmentation, machine translation, and text generation.[1,12–20]

Despite the success of these advanced methods, there remains a noticeable gap between their sophistication and the practical use by downstream users, particularly biomedical researchers and healthcare professionals. The technical intricacies represent significant burdens for them to directly apply those methods; this is particularly true for those lacking a background in computational methods or basic programming skills. Consequently, there is a growing demand for user-friendly and accessible toolkits designed to simplify medical text processing.

Multiple toolkits are available for text processing in the biomedical domain. Table 1 summarizes representative tools. While there are many other useful tools, here we mainly limit our comparison with Python-based open-source toolkits.

|  | ⋆ Question Answering | Text Summarization | Text Simplification | Machine Translation | Basic NLP Functions | Query Search |
|---|---|---|---|---|---|---|
| MIMIC-Extract[21] |  |  |  |  |  | ✓ |
| scispaCy[22] |  |  |  |  | ✓ |  |
| MedspaCy[23] |  |  |  |  | ✓ |  |
| Transformers-sklearn[24] |  |  |  |  | ✓ |  |
| Stanza Biomed[25] |  |  |  |  | ✓ |  |
| **Ascle** | ✓ | ✓ | ✓ | ✓ | ✓ | ✓ |

**Table 1.** A comparison with existing toolkits. The ⋆ denotes the task conducted human evaluation. Basic NLP Functions include abbreviation extraction, sentence tokenization, word tokenization, negation detection, hyponym detection, UMLS concept extraction, named entity recognition, document clustering, POS tagging, entity linking, text summarization (extractive methods) and multi-choice QA. It is worth noting that not every toolkit includes these 12 basic NLP functions, but Ascle includes them all.

MIMIC-Extract[21] is a pipeline for data extraction, preprocessing and representation from MIMIC-III dataset; scispaCy[22] is a tool that adapts spaCy's models for processing scientific and biomedical text; MedspaCy[23], also based on the spaCy framework, provides both rule-based and machine learning-based methods for processing medical text; Transformers-sklearn[24] enables the seamless integration of pre-trained Transformer-based models into the scikit-learn framework; Stanza Biomed[25] is a more advanced tool for statistical, neural, and rule-based problems in computational linguistics, and it provides a simple interface for NLP tasks with nearly state-of-the-art performance using neural networks. Additionally, there are toolkits such as CLAMP[26], which is designed for clinical text extraction and offers advanced NLP components along with a user-friendly graphical interface; cTAKES[27], specialized in extracting information from electronic medical record clinical free-text; and MetaMap[28], known for identifying UMLS concepts in text. However, these existing toolkits tend to emphasize different perspectives, and the absence of generation capabilities in any of them leaves a significant void.

In response, we present Ascle, a groundbreaking NLP toolkit specialized in medical text generation, which for the first time, includes four advanced generative functions: question-answering, text summarization, text simplification, and machine translation. Ascle also comprises 12 basic NLP functions; as well as query and search capabilities.[29] In addition, we fine-tuned 32 domain-specific language models, evaluated them thoroughly on 24 established benchmarks and conducted manual reviews with two healthcare professionals. Ascle empowers a diverse spectrum of users, from novices to experienced professionals, enabling them to effortlessly address their NLP tasks, even with limited technical expertise in handling textual data. We believe that Ascle not only democratizes access to cutting-edge methods but also expedites their integration into healthcare.

## MATERIALS AND METHODS

Ascle is a pioneering NLP toolkit designed specifically for medical text generation, which consists of three modules. Generative Functions is the core module of Ascle, which includes four advanced generative tasks: question-answering, text summarization, text simplification, and machine translation, covering various application scenarios in healthcare. Additionally, Ascle integrates 12 essential NLP functions, along with query and search capabilities for clinical databases. The overall architecture of Ascle is shown in Figure 1. This section will focus on introducing the core module of Ascle - Generative Functions. For more information on basic NLP functions and query and search module within Ascle, please refer to the discussion section.

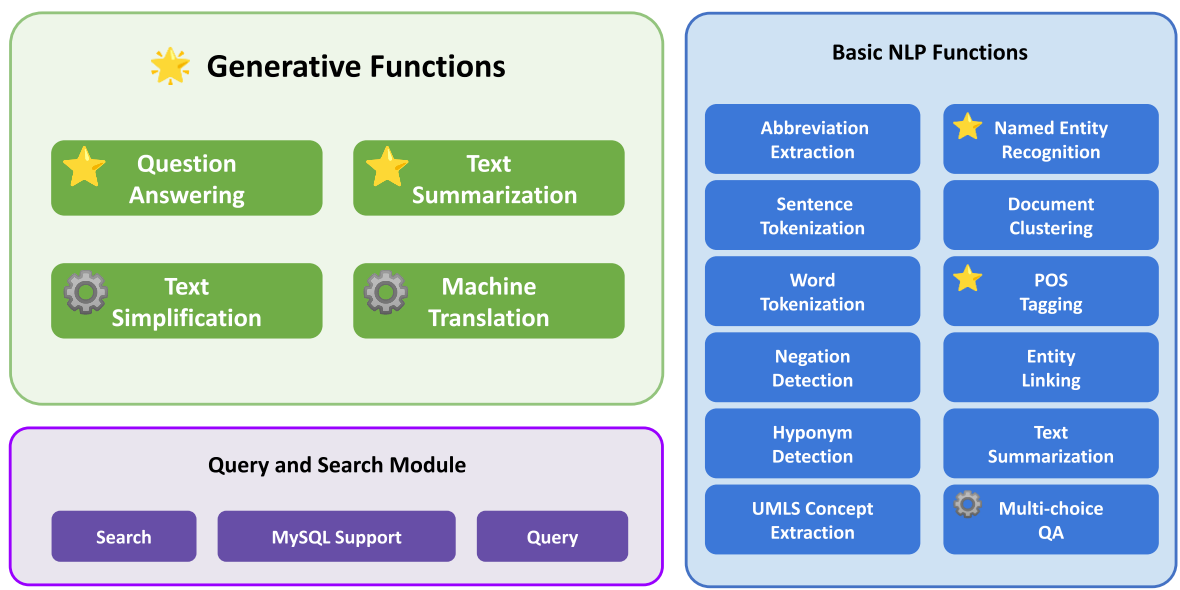

**Figure 1.** The overall architecture of Ascle. ⚙ indicates that we have our fine-tuned models for this task. ☆ indicates that we conducted evaluations for this task.

**Generative functions**

Ascle utilizes pre-trained language models to offer a range of generative functions including question-answering, text summarization, text simplification, and machine translation. It also grants users the flexibility to access any publicly available language models. Additionally, we provide fine-tuned models for specific generative tasks. All these fine-tuned models are publicly available for users to reference and utilize. In the following sections, we will introduce these powerful generative functions separately.

Question-answering

Question-answering is particularly crucial in healthcare.[30] When integrated into healthcare systems, it assumes roles such as pre-consultation and remote consultation, effectively coping with the exponential increase in patient load and alleviating the strain on the healthcare system. Moreover, specialized question-answering systems hold the potential to contribute to patient education and medical education.[19] In Ascle, we incorporated the question-answering function, which encompasses two sub-tasks: multiple-choice question-answering and answer generation.

*Multiple-choice question-answering*

We provide users with a biomedical multiple-choice question-answering function, allowing them to input question text and options to determine the most probable answer. While the primary approach for doing this entails a classification methodology, we have utilized generative models in the capacity of encoders. Consequently, we categorized this task here. We conducted comparative analysis on five biomedical and clinical pre-trained language models, including BioBERT, ClinicalBERT, SapBERT,[31] GatorTron-base,[32] and PubMedBERT. These models were fine-tuned and evaluated on the Head-QA[33] and MedMCQA datasets.[34] HEAD-QA covers six topics: medicine, nursing, psychology, chemistry, pharmacology, and biology, and all questions are sourced from professional position exams within the Spanish healthcare system. And MedMCQA is a larger dataset covering 2,400 healthcare topics and 21 medical subjects. Due to the lack of labels in the MedMCQA test set, we utilized the validation set for evaluation.

*Answer generation*

Apart from the multiple-choice question-answering task, we also provide the function of answer generation. For this task, we applied Baize-healthcare[35] and OPT-MedQuAD,[36] both of which were pre-trained on the MedQUAD[37] dataset. MedQUAD includes 47,457 medical question-answer pairs created from 12 National Institutes of Health (NIH) websites. We conducted evaluations using the QA Test Collection from the TREC-2017 LiveQA medical task,[38] which consists of 2,479 questions along with their corresponding reference answers. It's noteworthy that since objective metrics have been shown to be inaccurate in assessing the quality of generated content, we additionally conducted a manual validation in later sections. Two healthcare professionals carried out a manual review on 50 randomly sampled answers. The detailed evaluation guidance can be found in Supplementary Appendix A.

Text summarization

In healthcare, clinicians and researchers are confronted with an exponential surge of information, including literature, EHRs, and more.[39] Text summarization is an important generative task, aiming to distill essential information from the overwhelming complexity of texts and compress it into a more concise format.[40] Through automatic text summarization, clinicians and researchers can efficiently acquire information, thereby avoiding information overload.

We provide an abstractive text summarization function in this module, and compared general pre-trained summarization models including Pegasus,[41] BigBird,[42] BART,[43] PRIMERA,[44] which are pre-trained and fine-tuned on general text summarization corpora. We also compared domain-specific models such as SciFive[45] and BioBART,[46] which make use of biomedical corpora like Pubmed and PMC. Furthermore, we chose PubMed,[47] MIMIC-CXR,[48] and MEDQA-AnS[49] datasets for evaluation. The PubMed dataset consists of 133k biomedical scientific publications from the PubMed database. Each input document is a scientific article, and the reference summarization is the associated abstract. MIMIC-CXR is a de-identified, protected health information removed dataset of chest radiographs, with a DICOM format and free-text radiology reports. We used a subset from the MIMIC-CXR for the MEDIQA 2021 Radiology report summarization shared task.[50] Since we were unable to obtain the test set, we applied the validation set as the test set and additionally extracted 2000 instances from the training set to form a new validation set. MEDQA-AnS is a collection of consumer health questions and passages that contain information relevant to the question. It supports both single-document and multiple-document summarization evaluation.

Text simplification

Biomedical texts are typically laden with intricate terminologies, which can hinder the understanding of individuals without a clinical background.[51] In Ascle, the function for text simplification is to translate complex and technical biomedical texts into understandable content. This will enhance the comprehension and involvement of non-clinical individuals, including patients, enabling them to better engage with the information and participate in clinical decisions more effectively.

We evaluated several pre-trained models including BigBirdPegasus,[42] BART, and BioBART on the eLife, PLOS[52] and MedLane[53] datasets. The eLife and PLOS are shared task data released from the BioLaySumm 2023 Task 1, and the task aims to generate lay summarization given longer inputs. While eLife and PLOS are from the shared task, we did not obtain the ground truth of the original test set. To have a fair comparison, we conducted testing on the development dataset and left out some examples from the original training set for validation. MedLane is a large-scale human-annotated dataset containing professional-to-customer sentences selected from MIMIC-III. For MedLane, we split 2,030 examples from the training set as the validation set and used the original test set for evaluation. We fine-tuned on selected pre-trained models including Pegasus, BART, and BioBART.

Machine translation

Language barriers pose difficulties for patients to access timely information and communicate effectively with healthcare providers, resulting in low-quality healthcare services.[54] Our machine translation function aims to translate the text from a source language into a target language in a clinical scenario. Taking advantage of pre-trained models, Ascle supports 17 languages. We fine-tuned the existing MarianMT[55] and multilingual T5[56] using UFAL Medical Corpus which includes various medical text sources, such as titles of medical Wikipedia articles, medical term-pairs, patents, and documents from the European Medicines Agency. During the preprocessing phase, we excluded general domain data from UFAL, such as parliamentary proceedings, and randomly shuffled the medical-domain corpora, splitting them into two parts at a ratio of 85% and 15% for training and testing, respectively. For each language pair, we utilized all available parallel data to maximize the breadth and accuracy of our machine translation function.

# RESULTS

## Question-answering

Multiple-choice question-answering

We employed five biomedical pre-trained models for fine-tuning: BioBERT, ClinicalBERT, SapBERT, GatorTron-base, PubMedBERT, and utilized accuracy score as the evaluation metric, as shown in Figure 2. The findings reveal that among these, PubMedBERT excels on HEAD-QA and MedMCQA (without context) with accuracy rates of 42.52% and 46.59% respectively. Conversely, SapBERT, PubMedBERT and GatorTron-base achieve a very similar performance on MedMCQA (with context), especially GatorTron-base emerges as the superior performer, boasting an accuracy of 64.93%.

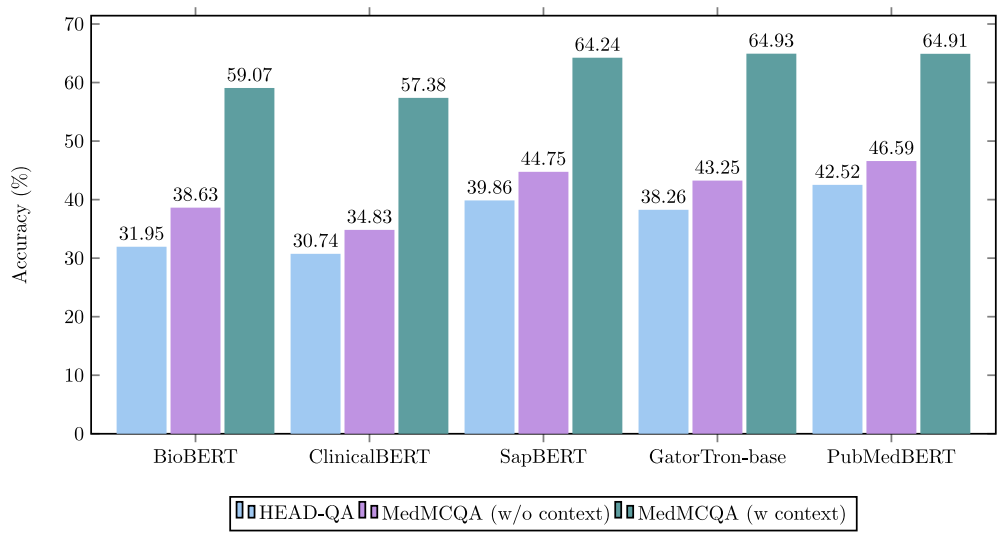

**Figure 2.** Evaluation for multiple-choice question-answering task.

Answer generation

We utilized ROUGE scores[57] to evaluate the answer generation capabilities of two pre-trained models: Baize-healthcare and OPT-MedQuAD. Baize-healthcare outperforms OPT-MedQuAD on all R-1, R-2, and R-L scores, with scores of 21.11, 5.14, 19.27 respectively. However, as previously noted, the metrics we used fall short in fully assessing the quality of content generated in healthcare. To address this gap, two healthcare professionals conducted manual reviews from four perspectives: Readability, Relevancy, Accuracy, and Completeness. We provide detailed results in the manual validation section.

## Text summarization

We evaluated text summarization on both single-document and multi-document settings, using ROUGE scores. In Table 2, we compared five selected models on four chosen benchmarks for the single-document scenario. For a fair comparison, we excluded the results of BioBART and SciFive, as they were fine-tuned on PubMed. We can observe that BART has a steady good performance on three benchmarks. Notably, BART demonstrates consistently strong performance on three of the benchmarks. An observation is that SciFive lags behind both BART and BioBART in terms of competitiveness. Additionally, BioBART only outperforms BART on one of the benchmarks.

|  | PubMed | | | MIMIC-CXR | | | MEDQA-AnS (p) | | | MEDQA-AnS (s) | | |
|---|---|---|---|---|---|---|---|---|---|---|---|---|
|  | R-1 | R-2 | R-L | R-1 | R-2 | R-L | R-1 | R-2 | R-L | R-1 | R-2 | R-L |
| Pegasus | 45.97 | 20.15 | 28.25 | 22.49 | 11.57 | 20.35 | 18.29 | 4.82 | 13.87 | 22.21 | 8.23 | 16.76 |
| BigBird | 46.32 | 20.65 | **42.33** | 38.99 | 29.52 | 38.59 | 13.18 | 2.14 | 10.04 | 14.89 | 3.13 | 11.15 |
| BART | **48.35** | **21.43** | 36.90 | **41.70** | **32.93** | **41.16** | **24.02** | **7.20** | **17.09** | 38.19 | 22.20 | 30.58 |
| SciFive | - | - | - | 35.41 | 26.48 | 35.07 | 13.08 | 2.15 | 10.10 | 16.88 | 6.47 | 14.42 |
| BioBARt | - | - | - | 41.61 | 32.90 | 41.00 | 22.58 | 7.49 | 16.69 | **39.40** | **24.64** | **32.07** |

**Table 2.** Evaluation for single-document summarization. Some results are derived from other papers.[58]

Furthermore, our evaluation extends to multi-document summarization using the MEDQA-AnS dataset, as shown in Table 3. We compared various models, including both traditional and deep learning approaches. For extractive summarization, we employ TextRank,[59] while for abstractive summarization, we consider BART, Pegasus, PRIMERA, and BioBART. Notably, BART demonstrates competitive performance, whereas BioBART exhibits slightly inferior results.

|  | MEDQA-AnS (p) | | | MEDQA-AnS (s) | | |
|---|---|---|---|---|---|---|
|  | R-1 | R-2 | R-L | R-1 | R-2 | R-L |
| TextRank | 29.88 | 10.23 | 17.01 | 43.77 | 26.80 | 30.52 |
| BART | **24.56** | **7.56** | **17.18** | **32.32** | 15.42 | **24.03** |
| Pegasus | 17.44 | 5.36 | 13.44 | 19.54 | 7.46 | 14.93 |
| PRIMERA | 16.66 | 4.89 | 12.68 | 21.78 | 9.77 | 16.85 |
| BioBARt | 23.16 | 7.47 | 16.47 | 30.87 | **15.91** | 23.66 |

**Table 3.** Evaluation for multi-document summarization.

**Text simplification**

We compared BigBirdPegasus, BART, and BioBART, and fine-tuned them for text simplification tasks. We used ROUGE scores for evaluation, as shown in Figure 3 (A). Interestingly, both BART and BioBART outperformed BigBirdPegasus across all three datasets. While BioBART, pre-trained on biomedical corpora on top of BART, demonstrates a slightly better performance only on a single dataset.

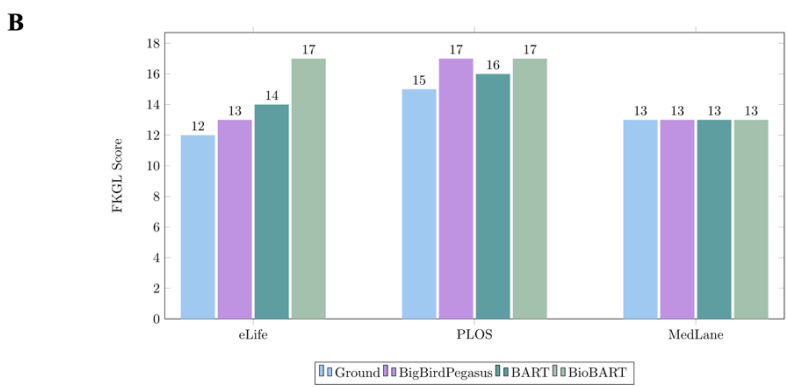

**Figure 3.** (A) Evaluation for text simplification task using ROUGE scores. (B) Evaluation for text simplification task using FKGL score.

Furthermore, we conducted an analysis of reading ability using the Flesch-Kincaid grade level (FKGL) score.[60] The FKGL score is a measure of text complexity and indicates the difficulty of understanding a given text, as shown in Figure 3 (B). We compared the outputs generated by our models with the ground truth. For the eLife and PLOS datasets, the ground truth exhibits FKGL scores of 12 and 15, respectively. Interestingly, the BioBART model performs competitively in terms of ROUGE metrics but fails to significantly reduce the difficulty of understanding, as evidenced by its FKGL score of 17 in both datasets. On the other hand, the BART model manages to slightly lower the FKGL score to 14 and 16 for eLife and PLOS, respectively. However, in the case of the MedLane dataset, all methods appear to reach a similar level of complexity as the ground truth. This can be attributed to the dataset's shorter examples and potentially smaller vocabulary size, which limits the observed differences.

**Machine translation**

We fine-tuned MarianMT and mT5 on three language pairs: "en-es", "en-fr", "en-ro", and used MarianMT as the baseline for comparison. BLEU score[61] was utilized for evaluation, as shown in figure 4. After fine-tuning, the BLEU scores significantly improved, with the most substantial improvement observed in the "en-fr" language pair. This enhancement can be attributed to the larger amount of training data available for "en-fr" (2,812,305 samples). Furthermore, across all three language pairs, the mT5 model outperformed the MarianMT model in terms of BLEU scores. We also fine-tuned mT5 on five language pairs: "en-cs", "en-de", "en-hu", "en-pl" and "en-sv"; exact results can be viewed in the Supplementary Appendix B.

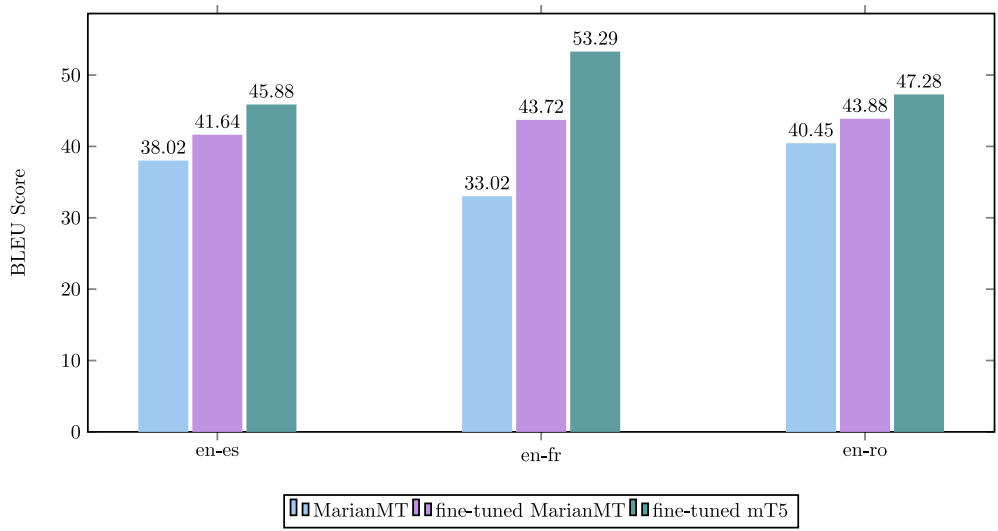

**Figure 4.** Evaluation for machine translation task.

## DISCUSSION
### Manual validation
We performed manual validation on the answer generation task. 50 question-answer pairs from QA Test Collection were randomly selected, with answers generated by Baize-healthcare. Subsequently, two healthcare professionals (one resident and one attending specialist) rated these generated answers on the criteria of Readability, Relevancy, Accuracy, and Completeness, using a 5-point Likert scale. Figure 5 (A) displays the average scores.

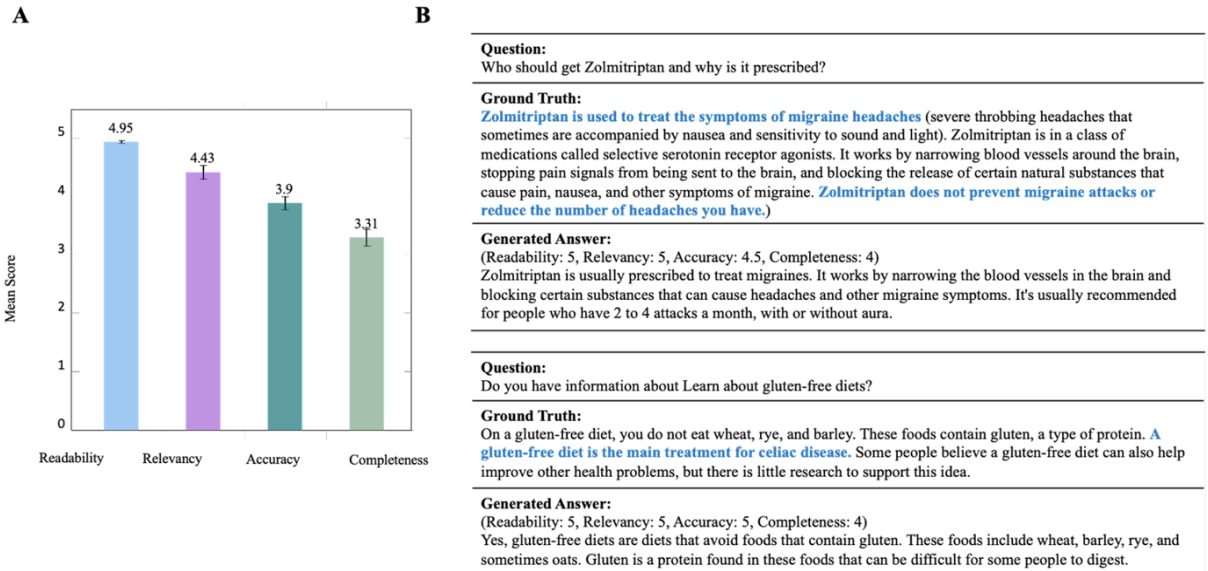

**Figure 5.** (A) Manual evaluation (Readability, Relevancy, Accuracy, Completeness) for 50 question-answer pairs. (B) Two examples of generated answers with ground truth.

The generated answers have good readability and relatively good relevancy, with scores of 4.95 and 4.43, respectively. In contrast, the completeness score is relatively lower (3.31). We observe that the generated answers may lack a comprehensive explanation. Figure 5 (B) presents two examples in detail. In the first example, compared to the ground truth, the generated answer does not point out that Zolmitriptan is used for treating acute migraines, nor does it indicate that it cannot be used to prevent migraine attacks or to reduce the frequency of headaches. And in the second example, the generated answer does not mention that a gluten-free diet is the main treatment for celiac disease. Additionally, we provide detailed presentations of two additional cases in Supplementary Appendix C.

Additionally, we calculated the Inter-evaluator Agreement (IAA) using percentage agreement for each criterion. Due to limitations in the number of questions and metrics, we categorized the scores into two groups: 0 for scores below 3, and 1 for scores 3 and above. Two healthcare professionals demonstrated a high level of consistency across all criteria, with the percentage agreement consistently exceeding 0.6. This is especially true for readability and relevance, which show minimal disparity. The consistency in accuracy and completeness slightly decreased, reaching 0.68 and 0.74, respectively, but these scores are still acceptable.

**Analysis of text summarization**

In the task of multi-document summarization, we included models based on traditional methods such as TextRank, as well as deep learning-based generative models like BART, Pegasus, PRIMERA, and BioBART. We evaluated their performance using ROUGE scores. However, it is noteworthy that despite TextRank's outperforms almost all generative models in ROUGE scores, this does not necessarily indicate superior performance. ROUGE scores are calculated based on the overlap between the generated content and reference summary. Given that TextRank is an extractive summarization model, it tends to score higher by this measure.

While generative models can distill complex information into an easy-to-understand format, demonstrating semantic comprehension abilities. As shown in Table 4, the summarizations generated by BART display well-structured patient information, with a brief description of events and corresponding conditions of the current patient (highlighted in blue), exhibiting high readability. In contrast, the summarizations produced by TextRank are less readable and include noise (highlighted in orange); the generated content is often a literal collage of text fragments. Despite TextRank achieving higher ROUGE scores, it lacks the ability to discern information and integrate it into coherent and readable content, showing significant limitations for practical use.

| BART | TexTRank |
|---|---|
| **The patient is a XXX-year-old man** with a history of a question of coronary artery disease, borderline diabetes mellitus. **He was in his usual state of health** until 11 p.m. last night when he experienced chest pain with radiation to his back, positive shortness of breath, positive diaphoresis, no vomiting, no lightheadedness. The patient had had a similar episode of chest pain and was taken to a XXX. **He had successful angioplasty and stent of LAD and CX**. **He is a middle aged XXX man** in no acute hypertensivedistress. He has had anginal chest pain which is similar to his presenting complaint, but without radiations to hisBack. His blood pressure was 105/73, pulse 84, respiratory 21, O2 saturation 92% on 2 liters. His CPK was 594, The index was 7.7, and he was admitted to the hospital with a high blood pressure. His condition was described as "stable" and "normal" by the doctor. The doctor referred the patient to a cardiologist for further treatment. The cardiologist said the patient was in good condition and should be discharged in a few days. | Admission Date: XXX    Discharge Date: XXXDate of Birth: XXX    Sex: MService: CCU-6HISTORY OF PRESENT ILLNESS:  The patient is a XXX-year-old manwith a history of a question of coronary artery disease,status post myocardial infarction in [**December 2175**],hypertension, borderline diabetes mellitus who was in hisusual state of health until 11 p.m. last night when, while[**4-12**] midsternal pressure like chest pain with radiation toback, positive shortness of breath, positive diaphoresis,positive nausea, no vomiting, no lightheadedness. Mucous membranes moist.Oropharynx clear.NECK: No jugular venous distention, no carotid bruits.CARDIOVASCULAR:  Regular rate, S1, S2, artificial S1 gallopand balloon pump, no murmurs or rubs.LUNGS:  Bibasilar rales, left greater than right.ABDOMEN:  Normoactive bowel sounds, nontender, nondistended.EXTREMITIES:  No cyanosis, clubbing or edema.NEUROLOGIC:  Alert and oriented x3.LABS AT OUTSIDE HOSPITAL: CPK was 304, troponin 1.75.Electrocardiogram at 1:23 a.m. was normal sinus rhythm at101, normal axis deviation, **2 to [**Street Address(2) 1755**]** elevation V1 to V5,Q V3, AVF.LABS AT [**Hospital6 **] AT 8 A.M.:  CBC- white blood cells 11.2, hemoglobin 13.0, hematocrit 36.7,platelets 232. CARDIOVASCULAR:  Coronary artery disease: Three vesseldisease with successful intervention on LAD and leftcircumflex, but RCA not done secondary to good collateral.The patient was continued on aspirin 325 qd. |
| **Patient has** CABG complicated by postop bleed and pleural effusion with discharge to [**Hospital1 **] Rehabilitation presents with abdominal pain. Zosyn was given in the ED. **Patient was otherwise doing well and was to go back to** rehab to finish his course of Cipro and Flagyl on [**5-17**]. Patient was last seen normal sometime last evening. **He woke up and noticed that the left side of his body felt "numb". He was not aware of any other** neurologicalweakness, and mostly complained of being very tired. He denied any new vision problems, did not have a headache. He sounded somewhat slurred but did not feel as if hisspeech was changed significantly. **He felt sleepy** but able to sustain attention, currently apparently in no distress. He was on standing.Plavix and [**State **] which had been | Of note he was on standingPlavix and [**State **] which had been held for the last few days (atleast since the 14), since he had the percutaneous drainage.The patient was otherwise doing well and was to go back to rehabto finish his course of Cipro and Flagyl on [**5-17**].Past Medical History:coronary artery disease s/p right coronary artery stent x2([**10-3**], [**3-4**]), hypertension, hyperlipidemia, chronic obstructivepulmonary disease, asbestos exposure, chronic back pain,insomnia and obstructive sleep apnea (untreated)PSH:[**2144-4-21**]Endoscopic, minimally invasive, off pump coronary artery bypassgraft x1 with left internal mammary artery to left anteriordescending artery.[**2144-4-21**]Re-exploration for bleeding, post coronary artery |
| held for the last few days (atleast since the 14), since he had the percutaneous drainage. He did not. feel that the weakness had progress and reported that he felt the sense of numbness was starting to improve and had some difficulty squeezing an examiners hand. He is a retired postal worker. He lives with wife. and son who is a chiropractor. | bypassgrafting.Social History:Lives with wife.Exposure to asbestos.**Defers all medical decisions** to son who is a chiropractor.Occupation: retired postal worker.Tobacco: 3 PPD x 30 years, quit 45 years agoETOH: NoneFamily History:Non-contributory to cholecystitis.Physical Exam:Physical Exam:Vitals: T: 97.9  P:75  R: 16  BP:128/73 SaO2:96**General: Awake, felt sleepy but able to sustain attention, poorhistorian currently.** |

**Table 4.** Two MIMIC-III (parts) examples of text summarization task, generated by BART and TextRank, respectively. (We eliminated sensitive information).

### Basic NLP functions

The basic NLP functions module in Ascle integrates many third-party libraries and supports up to 12 functions, including abbreviation extraction, sentence tokenization, word tokenization, negation detection, hyponym detection, UMLS concept extraction, named entity recognition, document clustering, POS tagging, entity linking, text summarization (extractive method) and multi-choice QA. Detailed information can be found in Supplementary Appendix D. We also selected POS tagging and named entity recognition tasks for evaluation, with the detailed results provided in the Supplementary Appendix E.

### Query and search capabilities

Ascle provides user-friendly query and search functions on clinical text corpora: (1) *MySQL Support for MIMIC-III database*: The data tables (i.e., NOTEEVENTS.TSV) were indexed into a MySQL database, and user-friendly interfaces were provided for basic statistical functions, such as obtaining the count of patients, documents, and sentences. (2) *Query*: We implemented a range of straightforward query functions. For instance, users could retrieve a specified number of patient records or notes by using their respective IDs. (3) *Search*: The effectiveness of search functionality within unstructured text was of paramount importance. To address this, we integrated keyword search capabilities supported by multiple libraries, thus enabling swift and targeted searches.

### System usage

Ascle provides an easy-to-use approach for biomedical researchers and healthcare professionals. Users can efficiently utilize it by merely inputting text and calling the required functions. Figure 6 displays two use cases.

```python
# create Ascle
from Ascle import Ascle
med = Ascle()

# Text Simplification
main_record = """
            The patient presents with symptoms of acute bronchitis,
            including cough, chest congestion, and mild fever.
            Auscultation reveals coarse breath sounds and occasional
            wheezing. Based on the clinical examination, a diagnosis
            of acute bronchitis is made, and the patient is prescribed
            a short course of bronchodilators and advised to rest and
            stay hydrated.
            """

# choose the model
layman_model = "ireneli1024/bart-large-elife-finetuned"

med.update_and_delete_main_record(content)

# call the text simplification function and print the output
print(med.get_layman_text(layman_model, min_length=20, max_length=70))

>> The patient presents with symptoms of acute bronchitis including
   cough, chest congestion and mild fever. Auscultation reveals coarse
   breath sounds and occasional wheezing. Based on these symptoms and
   the patient's history of previous infections with the same condition,
   the doctor decides that the patient is likely to have a cold or bronch.

# Machine Translation
main_record = """
            Myeloid derived suppressor cells (MDSC) are immature myeloid
            cells with immunosuppressive activity. They accumulate in
            tumor-bearing mice and humans with different types of cancer,
            including hepatocellular carcinoma (HCC).
            """

med.update_and_delete_main_record(record)

# call the machine translation function and print the output
print(med.get_translation_mt5("French"))

>> Les cellules suppressives dérivées de myéloïdes (MDSC) sont des
   cellules myéloïdes immatures ayant une activité immunosuppressive,
   accumulées chez des souris et des humains ayant différents types de
   cancer, y compris le carcinome hépatocellulaire (HCC).
```

**Figure 6.** Demonstration of system usage. We show two cases: text simplification and machine translation.

We also provide a detailed demonstration tutorial online:
https://github.com/Yale-LILY/Ascle/blob/v2.2/Ascle/demo.py
The documentation can be found in:
https://github.com/Yale-LILY/Ascle/blob/v2.2/Ascle/documentation.md

**Limitations and future work**

Evaluation Metrics

In the case of generation tasks, we primarily chose automatic metrics for evaluation, such as ROUGE and BLEU scores. However, it should be noted that existing automatic evaluation metrics, which calculate the overlap between generated content and reference are not suitable for healthcare scenarios. It has been shown that these metrics cannot effectively assess factual correctness[62] and may not align with human preference.[63] While human evaluation serves as an invaluable aspect in assessing the performance of the model, its incorporation may pose certain challenges due to various factors, including budget constraints.

Generalization

The primary objective of this toolkit is to offer user-friendly functions, providing convenience for healthcare professionals and researchers in processing medical texts. We incorporated and fine-tuned state-of-the-art language models for various tasks. However, it is important to acknowledge that the performance of these models might be constrained by the scale of training or pre-training data. Consequently, ensuring consistent performance when applied to user-provided data could be challenging, as the models may have limited generalization ability in such scenarios.

The extension to LLMs

In this work, Ascle focuses on fine-tuned domain-specific language models. In contrast, recent LLMs have shown great potential in generative applications especially its superior zero- and few-shot performance.[64–67] While LLMs show potential in biomedical and clinical applications,[30,68–71] studies have consistently demonstrated that when LLMs are applied in the biomedical and clinical field, the generated content can be unfaithful, inconsistent, incomplete and racially biased.[72–76] For example, when using LLMs for simplifying radiology report texts, "DD" is an abbreviation for differential diagnosis, but it is incorrectly recognized by LLMs as the final diagnosis, which could potentially cause harm to the patient.[77] In the question-answering task, when asked, "How is the eGFR calculated?", LLMs attempt to justify race-based medicine with false assertions about Black people having different muscle mass and therefore higher creatinine levels.[72] Additionally, in the text summarization task, LLMs can generate content that contradicts the facts contained in the input data, and they may also produce content that cannot be verified from the input data. We plan to thoroughly evaluate LLMs and extend to Ascle in the future, ensuring that their application truly benefits biomedical researchers and healthcare professionals.

Future work

Ascle primarily concentrates on utilizing pre-trained language models for generative tasks. These models are significantly smaller and more cost-effective compared to LLMs, offering local and lower-cost solutions. However, we plan to thoroughly evaluate popular LLMs and integrate them into Ascle in the future. Specifically, we will develop an extended module for LLMs to facilitate more effective applications in healthcare, such as enhanced medical question-answering systems.[30,78] These systems will not only focus on delivering accurate and factual information but will also incorporate evidence-based reasoning to support their responses. More importantly, as a toolkit, we will continue making the interface easy to access and user-friendly, ensuring it caters to both technical and non-technical users in the medical field.

**CONCLUSION**

We introduce Ascle, a comprehensive and pioneering NLP toolkit designed specifically for medical text generation. For the first time, it integrates generative functions, including question-answering, text summarization, text simplification and machine translation. Moreover, multiple healthcare professionals conducted manual reviews of the answer generation task. Our research fills the gap of existing toolkits for generative tasks, which holds significant implications for the entire healthcare domain. Ascle boasts remarkable flexibility, allowing users to access a variety of cutting-edge pre-trained models. Meanwhile, it stands as a user-friendly toolkit, ensuring ease of use even for clinical staff without a technical background. We will continue to maintain and extend Ascle.

## FUNDING
Tiarnan D.L. Keenan and Emily Y Chew were supported by the NIH Intramural Research Program (IRP), National Eye Institute. Zhiyong Lu, and Qingyu Chen were supported by the NIH IRP, National Library of Medicine. Qingyu Chen was also supported by the National Library of Medicine of the National Institutes of Health under award number 1K99LM014024.

## AUTHOR CONTRIBUTIONS
Rui Yang, Qingcheng Zeng, Keen You, Yujie Qiao, Lucas Huang, Chia-Chun Hsieh, Benjamin Rosand, Jeremy Goldwasser and Irene Li performed the data collection, data processing and experiments. Amisha D Dave, Tiarnan D.L. Keenan conducted manual reviews. Rui Yang, Qingyu Chen, Irene Li created the figures and tables, and drafted the manuscript. Emily Y Chew, Dragomir Radev, Zhiyong Lu, Hua Xu, Qingyu Chen and Irene Li were responsible for project administration. All authors conceived of the idea for the article.

## CONFLICT OF INTEREST STATEMENT
The authors do not have conflicts of interest related to this study.

## DATA AVAILABILITY STATEMENT
The toolkit, its models, and associated data are publicly available:
https://github.com/Yale-LILY/Ascle.
We also provide a detailed demonstration tutorial online:
https://github.com/Yale-LILY/Ascle/blob/v2.2/Ascle/demo.py.
The documentation can be found in:
https://github.com/Yale-LILY/Ascle/blob/v2.2/Ascle/documentation.md.

## REFERENCES

1. Li I, Yasunaga M, Nuzumlalı MY, Caraballo C, Mahajan S, Krumholz H, *et al.* A Neural Topic-Attention Model for Medical Term Abbreviation Disambiguation 2019.
2. Neural Natural Language Processing for unstructured data in electronic health records: A review. *Computer Science Review* 2022;**46**:100511.
3. International Society for Biocuration. Biocuration: Distilling data into knowledge. *PLoS Biol* 2018;**16**:e2002846.
4. Shickel B, Tighe PJ, Bihorac A, Rashidi P. *Deep EHR: A Survey of Recent Advances in Deep Learning Techniques for Electronic Health Record (EHR) Analysis*. n.d. URL: https://ieeexplore.ieee.org/abstract/document/8086133 (Accessed 21 October 2023).
5. al-Aiad A, Duwairi R, Fraihat M. *Survey: Deep Learning Concepts and Techniques for Electronic Health Record*. n.d. URL: https://ieeexplore.ieee.org/abstract/document/8612827 (Accessed 21 October 2023).
6. Zhang Y, Chen Q, Yang Z, Lin H, Lu Z. BioWordVec, improving biomedical word embeddings with subword information and MeSH. *Scientific Data* 2019;**6**:1–9.
7. Chen Q, Peng Y, Lu Z. *BioSentVec: creating sentence embeddings for biomedical texts*. n.d. URL: https://ieeexplore.ieee.org/abstract/document/8904728 (Accessed 21 October 2023).
8. Devlin J, Chang M-W, Lee K, Toutanova K. BERT: Pre-training of Deep Bidirectional Transformers for Language Understanding 2018.
9. Lee J, Yoon W, Kim S, Kim D, Kim S, So CH, *et al.* BioBERT: a pre-trained biomedical language representation model for biomedical text mining. *Bioinformatics* 2019;**36**:1234–40.

10  Alsentzer E, Murphy JR, Boag W, Weng W-H, Jin D, Naumann T, *et al.* Publicly Available Clinical BERT Embeddings 2019.
11  Yu Gu Microsoft Research, Redmond, WA, Robert Tinn Microsoft Research, Redmond, WA, Hao Cheng Microsoft Research, Redmond, WA, Michael Lucas Microsoft Research, Redmond, WA, Naoto Usuyama Microsoft Research, Redmond, WA, Xiaodong Liu Microsoft Research, Redmond, WA, *et al.* Domain-Specific Language Model Pretraining for Biomedical Natural Language Processing. *ACM Transactions on Computing for Healthcare (HEALTH)* 2021. https://doi.org/10.1145/3458754.
12  Singhal K, Tu T, Gottweis J, Sayres R, Wulczyn E, Hou L, *et al.* Towards Expert-Level Medical Question Answering with Large Language Models 2023.
13  Zhou T, Cao P, Chen Y, Liu K, Zhao J, Niu K, *et al.* Automatic ICD Coding via Interactive Shared Representation Networks with Self-Distillation Mechanism.
14  Li F, Yu H. ICD Coding from Clinical Text Using Multi-Filter Residual Convolutional Neural Network. *AAAI* 2020;**34**:8180–7.
15  Song B, Li F, Liu Y, Zeng X. Deep learning methods for biomedical named entity recognition: a survey and qualitative comparison. *Brief Bioinform* 2021;**22**:bbab282.
16  Badjatiya P, Kurisinkel LJ, Gupta M, Varma V. Attention-Based Neural Text Segmentation. *Advances in Information Retrieval* 2018:180–93.
17  Wei-Hung Weng Massachusetts Institute of Technology, Cambridge, MA, USA, Yu-An Chung Massachusetts Institute of Technology, Cambridge, MA, USA, Peter Szolovits Massachusetts Institute of Technology, Cambridge, MA, USA. *Unsupervised Clinical Language Translation*. n.d. https://doi.org/10.1145/3292500.3330710.
18  Ben Abacha A, Demner-Fushman D. On the Summarization of Consumer Health Questions.
19  Yang R, Tan TF, Lu W, Thirunavukarasu AJ, Ting DSW, Liu N. Large language models in health care: Development, applications, and challenges. *Health Care Science* 2023. https://doi.org/10.1002/hcs2.61.
20  Li Y, Wehbe RM, Ahmad FS, Wang H, Luo Y. Clinical-Longformer and Clinical-BigBird: Transformers for long clinical sequences 2022.
21  Wang S, McDermott MBA, Chauhan G, Hughes MC, Naumann T, Ghassemi M. MIMIC-Extract: A Data Extraction, Preprocessing, and Representation Pipeline for MIMIC-III 2019. https://doi.org/10.1145/3368555.3384469.
22  Neumann M, King D, Beltagy I, Ammar W. ScispaCy: Fast and Robust Models for Biomedical Natural Language Processing 2019. https://doi.org/10.18653/v1/W19-5034.
23  Eyre H, Chapman AB, Peterson KS, Shi J, Alba PR, Jones MM, *et al.* Launching into clinical space with medspaCy: a new clinical text processing toolkit in Python. *AMIA Annu Symp Proc* 2021;**2021**:438.
24  Yang F, Wang X, Ma H, Li J. Transformers-sklearn: a toolkit for medical language understanding with transformer-based models. *BMC Med Inform Decis Mak* 2021;**21**:1–8.
25  Zhang Y, Zhang Y, Qi P, Manning CD, Langlotz CP. Biomedical and clinical English model packages for the Stanza Python NLP library. *J Am Med Inform Assoc* 2021;**28**:1892–9.
26  Soysal E, Wang J, Jiang M, Wu Y, Pakhomov S, Liu H, *et al.* CLAMP – a toolkit for efficiently building customized clinical natural language processing pipelines. *J Am Med Inform Assoc* 2017;**25**:331–6.
27  Savova GK, Masanz JJ, Ogren PV, Zheng J, Sohn S, Kipper-Schuler KC, *et al.* Mayo clinical Text Analysis and Knowledge Extraction System (cTAKES): architecture, component evaluation and applications. *J Am Med Inform Assoc* 2010;**17**:507–13.
28  *MetaMap*. n.d. URL: https://lhncbc.nlm.nih.gov/ii/tools/MetaMap.html (Accessed 1 December 2023).
29  Johnson AEW, Pollard TJ, Shen L, Lehman L-WH, Feng M, Ghassemi M, *et al.* MIMIC-III, a freely accessible critical care database. *Scientific Data* 2016;**3**:1–9.
30  Yang R, Marrese-Taylor E, Ke Y, Cheng L, Chen Q, Li I. Integrating UMLS Knowledge into Large Language Models for Medical Question Answering 2023.
31  Liu F, Shareghi E, Meng Z, Basaldella M, Collier N. Self-Alignment Pretraining for Biomedical Entity Representations 2020.
32  Yang X, Chen A, PourNejatian N, Shin HC, Smith KE, Parisien C, *et al.* GatorTron: A Large

33. Vilares D, Gómez-Rodríguez C. HEAD-QA: A Healthcare Dataset for Complex Reasoning 2019.
Clinical Language Model to Unlock Patient Information from Unstructured Electronic Health Records 2022.
34. Pal A, Umapathi LK, Sankarasubbu M. MedMCQA: A Large-Scale Multi-Subject Multi-Choice Dataset for Medical Domain Question Answering. Presented at the Conference on Health, Inference, and Learning.
35. Xu C, Guo D, Duan N, McAuley J. Baize: An Open-Source Chat Model with Parameter-Efficient Tuning on Self-Chat Data 2023.
36. Zhang S, Roller S, Goyal N, Artetxe M, Chen M, Chen S, *et al.* OPT: Open Pre-trained Transformer Language Models 2022.
37. Ben Abacha A, Demner-Fushman D. A question-entailment approach to question answering. *BMC Bioinformatics* 2019;**20**:1–23.
38. Ben Abacha A, Agichtein E, Pinter Y, Demner-Fushman D. Overview of the Medical Question Answering Task at TREC 2017 LiveQA. *TREC*. 2017. p. 1–12.
39. Text summarization in the biomedical domain: A systematic review of recent research. *J Biomed Inform* 2014;**52**:457–67.
40. Xie Q, Luo Z, Wang B, Ananiadou S. A Survey for Biomedical Text Summarization: From Pre-trained to Large Language Models 2023.
41. Zhang J, Zhao Y, Saleh M, Liu P. PEGASUS: Pre-Training with Extracted Gap-Sentences for Abstractive Summarization. Presented at the International Conference on Machine Learning.
42. Zaheer M, Guruganesh G, Dubey KA, Ainslie J, Alberti C, Ontanon S, *et al.* Big Bird: Transformers for Longer Sequences. *Adv Neural Inf Process Syst* 2020;**33**:17283–97.
43. Lewis M, Liu Y, Goyal N, Ghazvininejad M, Mohamed A, Levy O, *et al.* BART: Denoising Sequence-to-Sequence Pre-training for Natural Language Generation, Translation, and Comprehension 2019.
44. Xiao W, Beltagy I, Carenini G, Cohan A. PRIMERA: Pyramid-based Masked Sentence Pre-training for Multi-document Summarization 2021.
45. Phan LN, Anibal JT, Tran H, Chanana S, Bahadroglu E, Peltekian A, *et al.* SciFive: a text-to-text transformer model for biomedical literature 2021.
46. Yuan H, Yuan Z, Gan R, Zhang J, Xie Y, Yu S. BioBART: Pretraining and Evaluation of A Biomedical Generative Language Model 2022.
47. Cohan A, Dernoncourt F, Kim DS, Bui T, Kim S, Chang W, *et al.* A Discourse-Aware Attention Model for Abstractive Summarization of Long Documents 2018.
48. Johnson AEW, Pollard TJ, Berkowitz SJ, Greenbaum NR, Lungren MP, Deng C-Y, *et al.* MIMIC-CXR, a de-identified publicly available database of chest radiographs with free-text reports. *Scientific Data* 2019;**6**:1–8.
49. Savery M, Abacha AB, Gayen S, Demner-Fushman D. Question-driven summarization of answers to consumer health questions. *Scientific Data* 2020;**7**:1–9.
50. Delbrouck J-B, Zhang C, Rubin D. QIAI at MEDIQA 2021: Multimodal Radiology Report Summarization.
51. Devaraj A, Wallace BC, Marshall IJ, Li JJ. Paragraph-level Simplification of Medical Texts. *Proceedings of the Conference Association for Computational Linguistics North American Chapter Meeting* 2021;**2021**:4972.
52. Goldsack T, Zhang Z, Lin C, Scarton C. Making Science Simple: Corpora for the Lay Summarisation of Scientific Literature 2022.
53. Luo J, Zheng Z, Ye H, Ye M, Wang Y, You Q, *et al.* A Benchmark Dataset for Understandable Medical Language Translation 2020.
54. Khoong EC, Rodriguez JA. A Research Agenda for Using Machine Translation in Clinical Medicine. *J Gen Intern Med* 2022;**37**:1275–7.
55. Junczys-Dowmunt M, Grundkiewicz R, Dwojak T, Hoang H, Heafield K, Neckermann T, *et al.* Marian: Fast Neural Machine Translation in C++ 2018.
56. Xue L, Constant N, Roberts A, Kale M, Al-Rfou R, Siddhant A, *et al.* mT5: A massively multilingual pre-trained text-to-text transformer 2020.
57. Lin C-Y. ROUGE: A Package for Automatic Evaluation of Summaries.

58  Rohde T, Wu X, Liu Y. Hierarchical Learning for Generation with Long Source Sequences 2021.
59  Mihalcea R, Tarau P. TextRank: Bringing Order into Text.
60  Kher A, Johnson S, Griffith R. Readability Assessment of Online Patient Education Material on Congestive Heart Failure. *Adv Prev Med* 2017;**2017**:9780317.
61  Papineni K, Roukos S, Ward T, Zhu W-J. BLEU. Presented at the the 40th Annual Meeting, Philadelphia, Pennsylvania, 2002/7/7-2002/7/12.
62  Xie Q, Schenck EJ, Yang HS, Chen Y, Peng Y, Wang F. Faithful AI in Medicine: A Systematic Review with Large Language Models and Beyond. *medRxiv* 2023:2023.04.18.23288752–2023.04.18.23288752. https://doi.org/10.1101/2023.04.18.23288752.
63  Fleming SL, Lozano A, Haberkorn WJ, Jindal JA, Reis EP, Thapa R, *et al.* MedAlign: A Clinician-Generated Dataset for Instruction Following with Electronic Medical Records 2023.
64  Gao F, Jiang H, Blum M, Lu J, Liu D, Jiang Y, *et al.* Large Language Models on Wikipedia-Style Survey Generation: an Evaluation in NLP Concepts 2023.
65  Song L, Zhang J, Cheng L, Zhou P, Zhou T, Li I. NLPBench: Evaluating Large Language Models on Solving NLP Problems 2023.
66  Brown T, Mann B, Ryder N, Subbiah M, Kaplan JD, Dhariwal P, *et al.* Language Models are Few-Shot Learners. *Adv Neural Inf Process Syst* 2020;**33**:1877–901.
67  OpenAI. GPT-4 Technical Report 2023.
68  Liu S, McCoy AB, Wright AP, Carew B, Genkins JZ, Huang SS, *et al.* Leveraging Large Language Models for Generating Responses to Patient Messages. *medRxiv* 2023:2023.07.14.23292669. https://doi.org/10.1101/2023.07.14.23292669.
69  Wang H, Liu C, Xi N, Qiang Z, Zhao S, Qin B, *et al.* HuaTuo: Tuning LLaMA Model with Chinese Medical Knowledge 2023.
70  Wang G, Yang G, Du Z, Fan L, Li X. ClinicalGPT: Large Language Models Finetuned with Diverse Medical Data and Comprehensive Evaluation 2023.
71  Chen Z, Cano AH, Romanou A, Bonnet A, Matoba K, Salvi F, *et al.* MEDITRON-70B: Scaling Medical Pretraining for Large Language Models 2023.
72  Omiye JA, Lester JC, Spichak S, Rotemberg V, Daneshjou R. Large language models propagate race-based medicine. *Npj Digital Medicine* 2023;**6**:1–4.
73  He K, Mao R, Lin Q, Ruan Y, Lan X, Feng M, *et al.* A Survey of Large Language Models for Healthcare: from Data, Technology, and Applications to Accountability and Ethics 2023.
74  Tian S, Jin Q, Yeganova L, Lai P-T, Zhu Q, Chen X, *et al.* Opportunities and Challenges for ChatGPT and Large Language Models in Biomedicine and Health. *ArXiv* 2023.
75  Chen Q, Du J, Hu Y, Keloth VK, Peng X, Raja K, *et al.* Large language models in biomedical natural language processing: benchmarks, baselines, and recommendations 2023.
76  Thirunavukarasu AJ, Ting DSJ, Elangovan K, Gutierrez L, Tan TF, Ting DSW. Large language models in medicine. *Nat Med* 2023;**29**:1930–40.
77  Jeblick K, Schachtner B, Dexl J, Mittermeier A, Stüber AT, Topalis J, *et al.* ChatGPT makes medicine easy to swallow: an exploratory case study on simplified radiology reports. *Eur Radiol* 2023:1–9.
78  Malaviya C, Lee S, Chen S, Sieber E, Yatskar M, Roth D. ExpertQA: Expert-Curated Questions and Attributed Answers 2023.

**Supplementary Appendix**

**Supplementary Appendix A: Evaluation Guidance for Human Validation**

**Readability**
The quality of the answer text, ignoring the input.

1 (bad): The text is highly difficult to read, full of grammatical errors, and lacks coherence and clarity.

2: The text is somewhat difficult to read, and there are occasional grammatical errors. The coherence and clarity could be improved.

3: The text is moderately easy to read, but there are noticeable grammatical errors and some parts lack coherence and clarity.

4: The text is fairly easy to read, with only a few minor grammatical errors. Overall coherence and clarity are good, but there is room for improvement.

5 (good): The text is easy to read, well-structured, and flows naturally.

**Relevancy**
The pertinence of the answer to the posed question.

1 (bad): The answer is entirely off-topic and does not address the question at all.

2: The answer somewhat addresses the question but contains a significant amount of irrelevant information.

3: The answer is moderately relevant to the question but could be more focused.

4: The answer is mostly relevant with only minor deviations from the topic.

5 (good): The answer directly addresses the question and stays on topic throughout.

**Accuracy**
The correctness and truthfulness of the information provided in the answer.

1 (bad): The answer contains entirely incorrect or misleading information.

2: The answer contains several inaccuracies or misleading statements.

3: The answer is somewhat accurate but has noticeable errors.

4: The answer is mostly accurate with only minor errors.

5 (good): The answer is entirely accurate and trustworthy.

**Completeness**
The extent to which the answer covers all aspects of the question **(compared with the ground truth)**.

1 (bad): The answer barely touches on the topic and leaves out most of the necessary information.

2: The answer covers some aspects of the question but misses several key points.

3: The answer provides a moderate amount of information but could be more comprehensive.

4: The answer is fairly comprehensive but misses a few minor details.

5 (good): The answer thoroughly addresses all aspects of the question and leaves no stone unturned.

**Supplementary Appendix B: Machine Translation Evaluation on Other Five Language Pairs**

We also fine-tuned mT5 on other five language pairs: "en-cs", "en-de", "en-hu", "en-pl", "en-sv", and evaluated the performance using the BLEU score.

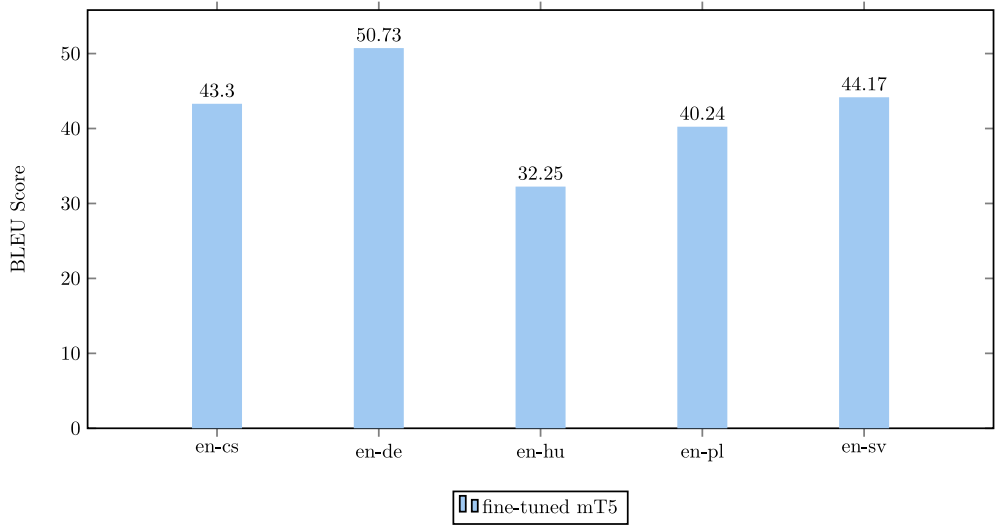

**Supplementary Appendix C: Case Study**

Based on the physicians' ratings, we selected two typical cases: In the first case, the generated answer is entirely consistent with the ground truth, achieving a score of 5 in readability, relevancy, accuracy, and completeness. In the second case, although the generated answer scored 5 in readability and scored 4 in relevancy, it is incorrect and completely contrary to the ground truth, receiving the lowest scores in accuracy and completeness.

| |
|---|
| **Question:** <br> What should I do if I forget a dose of Amphetamine ? |
| **Ground Truth:** <br> Take the missed dose as soon as you remember it. However, if it is almost time for the next dose, skip the missed dose and continue your regular dosing schedule. Do not take a double dose to make up for a missed one.) |
| **Generated Answer by Baize-healthcare** <br> **(Readability: 5, Relevancy: 5, Accuracy: 5, Completeness: 5)** <br><br> If you forget a dose of Amphetamine, you should take it as soon as you remember. However, if it is almost time for your next dose, skip the missed dose and continue with your regular dosing schedule. Do not take a double dose to make up for the missed one. |
| **Question:** <br> Is ring chromosome 20 syndrome inherited ? (Also called: r(20) syndrome; ring 20 syndrome; ring |

chromosome 20; ring chromosome 20 epilepsy syndrome)

**Ground Truth:**

**Ring chromosome 20 syndrome is almost never inherited.** A ring chromosome typically occurs as a random event during the formation of reproductive cells (eggs or sperm) or in early embryonic development. Often, the ring chromosome is present in only some of a person's cells. This situation is known as mosaicism. Most affected individuals have no history of the disorder in their families. However, at least one family has been reported in which a ring chromosome 20 was passed from a mother to her children.)

**Generated Answer by Baize-healthcare**
**(Readability: 5, Relevancy: 4, Accuracy: 1, Completeness: 1)**

**Yes, ring chromosome 20 syndrome is inherited in an autosomal recessive manner,** which means that a person must inherit two copies of the mutated gene (one from each parent) to develop the condition.

## Supplementary Appendix D: 12 Basic NLP Functions in Ascle

| Functions | Libraries |
|---|---|
| Abbreviation Extraction | scispaCy |
| Sentence Tokenization | MedspaCy, scispaCy, Stanza Biomed |
| Word Tokenization | MedspaCy, scispaCy, Stanza Biomed |
| Negation Detection | scispaCy |
| Hyponym Detection | scispaCy |
| UMLS Concept Extraction | scispaCy |
| Named Entity Recognition | MedspaCy, scispaCy, Stanza Biomed |
| Document Clustering | GloVe, Transformers |
| POS Tagging | MedspaCy, scispaCy, Stanza Biomed |
| Entity Linking | MedspaCy |
| Text Summarization | summa |
| Multi-choice QA | Transformers |

## Supplementary Appendix E: Evaluation of Basic NLP Functions

### POS Tagging

We conducted evaluations on 2019 CRAFT Shared Task (CRAFT-SA) and GENIA corpus. The testing case numbers are 9069 and 2036 respectively. We chose to test on various scispaCy models, a pretrained open-source model (flair/pos-english), and four Stanza Biomed models, with the results shown in the below figure. We concluded that scispaCy models are better on GENIA significantly, but on CRAFT, they all have a comparable performance.

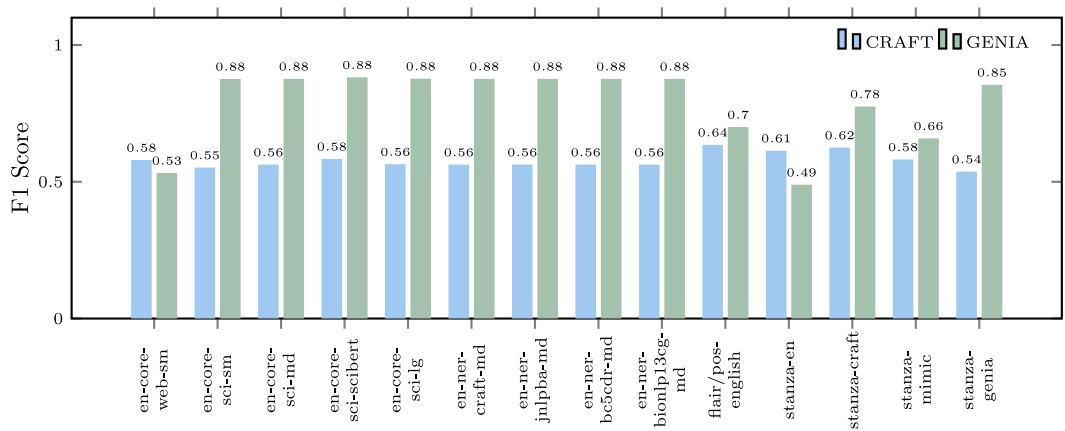

**Named Entity Recognition**

We chose NCBI disease corpus and BC5CDR (which contains both disease and chemical entity types) datasets for evaluation. The test sets for these two datasets consist of 941 and 4,797 instances, respectively. We compared seven different models from scispaCy and Stanza Biomed, as shown in below table. Stanza models exhibit superior performance, but it is noteworthy that these models were specifically pre-trained for these datasets, whereas in the scispaCy, only "scispaCy-bc5cdr-md" was specifically pre-trained on the BC5CDR dataset.

|  | NCBI-disease | | | BC5CDR-disease | | | BC5CDR-chem | | |
| --- | --- | --- | --- | --- | --- | --- | --- | --- | --- |
|  | R | P | F1 | R | P | F1 | R | P | F1 |
| scispaCy-sci-sm | 61.88 | 8.07 | 14.28 | 74.55 | 8.76 | 15.68 | 73.24 | 10.48 | 18.33 |
| scispaCy-sci-md | 62.50 | 7.98 | 14.16 | 76.60 | 8.89 | 15.94 | 77.51 | 10.96 | 19.20 |
| scispaCy-bc5cdr-md | 51.56 | 52.38 | 51.97 | 76.24 | 36.39 | 49.27 | 84.70 | 49.21 | 62.25 |
| Stanza-default | **83.44** | 82.83 | **83.13** | 76.51 | 78.10 | 77.30 | 80.58 | 86.01 | 83.20 |
| Stanza-mimic | 76.98 | **84.36** | 80.50 | 75.95 | 80.02 | 77.93 | 76.06 | 86.74 | 81.05 |
| Stanza-craft | 79.69 | 84.16 | 81.86 | **81.04** | 79.95 | 80.49 | **85.89** | 88.33 | **87.09** |
| Stanza-genia | 76.56 | 83.52 | 79.89 | 80.42 | **81.29** | **80.85** | 82.95 | **88.49** | 85.63 |